# BASIC: Semi-supervised Multi-organ Segmentation with Balanced Subclass Regularization and Semantic-conflict Penalty

Zhenghao Feng, Lu Wen, Yuanyuan Xu, Binyu Yan, Xi Wu, Jiliu Zhou, S*enior Member, IEEE*, Yan Wang, *Member, IEEE*

*Abstract*—Semi-supervised learning (SSL) has shown notable potential in relieving the heavy demand of dense prediction tasks on large-scale well-annotated datasets, especially for the challenging multi-organ segmentation (MoS). However, the prevailing class-imbalance problem in MoS caused by the substantial variations in organ size exacerbates the learning difficulty of the SSL network. To address this issue, in this paper, we propose an innovative semi-supervised network with **BA**lanced **S**ubclass regular**I**zation and semantic-**C**onflict penalty mechanism (BASIC) to effectively learn the unbiased knowledge for semi-supervised MoS. Concretely, we construct a novel auxiliary subclass segmentation (SCS) task based on priorly generated balanced subclasses, thus deeply excavating the unbiased information for the main MoS task with the fashion of multi-task learning. Additionally, based on a mean teacher framework, we elaborately design a balanced subclass regularization to utilize the teacher predictions of SCS task to supervise the student predictions of MoS task, thus effectively transferring unbiased knowledge to the MoS subnetwork and alleviating the influence of the class-imbalance problem. Considering the similar semantic information inside the subclasses and their corresponding original classes (i.e., parent classes), we devise a semantic-conflict penalty mechanism to give heavier punishments to the conflicting SCS predictions with wrong parent classes and provide a more accurate constraint to the MoS predictions. Extensive experiments conducted on two publicly available datasets, i.e., the WORD dataset and the MICCAI FLARE 2022 dataset, have verified the superior performance of our proposed BASIC compared to other state-of-the-art methods.

*Index Terms*— Semi-supervised Learning, Multi-organ Segmentation, Balanced Subclass Regularization, Mean Teacher

## I. INTRODUCTION

MULTI-ORGAN segmentation (MoS) is an imperative task in computer-assisted diagnosis (CAD) which aims to simultaneously assign an accurate class label to each pixel of multiple organs inside the medical images [1]. In the clinic, the organ boundaries are manually delineated by the physicians with a time-consuming process, posing an obstacle to timely follow-up treatment for patients. To ease the heavy burden on physicians and accelerate the delineation procedure, several deep learning (DL)-based methods have recently been proposed to automatically predict the organ contours and reached promising performance thanks to the fully supervised training on substantial labeled data [2, 3, 4]. However, collecting such a large amount of data with precise pixel-level annotation is unrealistic due to the expensive time and labor costs.

To reduce the reliance on extensive annotated data, semi-supervised learning (SSL), which simultaneously utilizes the limited labeled data and abundant unlabeled data to enhance the segmentation accuracy, has nowadays gained widespread research attention [5, 6, 7, 8, 9]. For example, to use the unlabeled data, the mean teacher [5] enforces the consistent predictions between a student and a teacher model where different data augmentations are adopted. Based on mean teacher, Hu *et al*. [6] introduced an attention mechanism for forcing the model to focus more on the region of interest (ROI), thus improving the segmentation accuracy for nasopharyngeal carcinoma (NPC) tumor. Furthermore, besides the segmentation task, Luo *et al*. [7] incorporated an additional regression task and harnessed a dual-task consistency regularization to further constrain the predictions. Nevertheless, these previous works mainly concentrate on the segmentation of a single ROI in a semi-supervised fashion which usually encounter unavoidable performance drops when directly applied to MoS task. Therefore, semi-supervised MoS is lately presented. However, until now, there have been only a few researches along the semi-supervised MoS research direction. Particularly, Zhou *et al*. [10] followed the idea of multi-view learning [11] and proposed a deep multi-planar co-training (DMPCT) model to predict the segmentation of multiple organs from different planar and gain the final output after a fusion operation. Nevertheless, they only explored a simple application of muti-view learning in semi-supervised MoS and not delved into its potential difficulties.

One of the most crucial problems in semi-supervised MoS is class imbalance raised by the substantial variations of the organ sizes where the classes with more observations (i.e., pixels) may overshadow the minority classes [12]. While the deep models try to simultaneously segment multiple organs, the intrinsic size variations among different organs bring about the models biasing towards the larger organs, thus degrading

This work is supported by National Natural Science Foundation of China (NSFC 62371325, 62071314), Sichuan Science and Technology Program 2023YFG0263, 2023YFG0025, 2023NSFSC0497. (Zhenghao Feng and Lu Wen contribute equally to this work; Yan Wang is the corresponding author.)

Zhenghao Feng, Lu Wen, Yuanyuan Xu, Binyu Yan, Jiliu Zhou, and Yan Wang are with the School of Computer Science, Sichuan University, China. (e-mail: fzh_scu@163.com; wenlu0416@163.com; YuanyuanXuSCU@outlook.com; yanby@scu.edu.cn; zhoujiliu@cuit.edu.cn; wangyanscu@hotmail.com).

Xi Wu is with the School of Computer Science, Chengdu University of Information Technology, China. (e-mail: wuxi@cuit.edu.cn).

the overall segmentation accuracy. Nowadays, several works have emerged in the fully supervised scenarios to rebalance the original data with multiple strategies. As an illustration, Tappeiner *et al.* [13] presented a class adaptive Dice loss to balance the penalties to different ROIs based on their pixel proportions. Bria *et al.* [14] designed a cascade of decision trees to drastically decrease the pixel number of large targets, thus handling the class imbalance issue. Nonetheless, these strategies encounter with following two limitations. First, these techniques mainly focus on fully supervised settings that rely on accurate labels to correct the biased predictions which are not applicable for the abundant unlabeled data in SSL scenarios. Second, such re-weighting or resampling methods lack further generation or utilization of the balanced data, resulting in limited performance enhancements. Consequently, it is essential to develop an appropriate solution to relieve the class imbalance problem in the semi-supervised MoS task.

In this paper, to address the above-mentioned issues, we present a novel semi-supervised network with **BA**lanced **S**ubclass regular**I**zation and semantic-**C**onflict penalty mechanism, namely BASIC, to learn unbiased knowledge for the challenging semi-supervised MoS task. Specifically, considering the performance of semi-supervised MoS heavily rely on the extraction of unbiased knowledge, we follow the idea of multi-task learning [15] and innovatively construct an auxiliary subclass segmentation (SCS) task along with the main MoS task targeting at balanced information learning. To achieve the SCS, we first use a class-balanced subclass partition strategy to produce multiple balanced subclasses from original biased classes (also called parent classes) according to their pixel proportions. Subsequently, the SCS and MoS tasks are fulfilled with a shared encoder and two task-specific decoders, thus strengthening the excavation the unbiased information. Then, motivated by the mean teacher [5] framework, we construct a teacher model and a student model and each contains both the two tasks. Additionally, we propose a novel balanced subclass regularization to constrain the MoS predictions of the student model with the SCS predictions of the teacher model, thus effectively transferring unbiased knowledge in the SCS subnetwork to the MoS subnetwork and mitigating the impact of the class-imbalance problem. Furthermore, considering the important hierarchical semantic relationship between the parent classes and subclasses (i.e., subclasses derived from the same parent class share similar semantic information), we elaborately devise a semantic-conflict penalty mechanism to provide heavier punishments to the conflict SCS predictions with wrong parent classes, thus giving a more precise constraint to the main MoS predictions. In summary, the main contributions of this work can be concluded as the following four-fold:

1) We propose a novel semi-supervised network, called BASIC, to simultaneously utilize the unlabeled data for the challenging semi-supervised MoS task and alleviate the negative impact of its inherent class imbalance issue.

2) We leverage the spirt of multi-task learning and construct an effective auxiliary SCS task based on balanced subclasses to provide the main MoS task with beneficial balanced information.

3) We present a novel balanced subclass regularization to additionally strengthen the transfer of unbiased knowledge into the MoS subnetwork, thus cleverly relieving the class-imbalance problem.

4) We design a semantic-conflict penalty mechanism to provide heavier punishments to the conflict SCS predictions with the wrong parent class and accomplish a more precise constraint to the main MoS predictions.

Extensive experiments are conducted on two public abdominal multi-organ segmentation datasets, i.e., the WORD dataset [16] as well as the MICCAI FLARE 2022 dataset [17], and the experimental results have confirmed the superior performance of the proposed BASIC network.

The remainder of this paper is structured as follows: In Section II, a brief review of related work is provided, the proposed network and its corresponding objective functions are described in Section III, the experimental setup and results are presented in Section IV, a comprehensive discussion of the proposed method is stated in Section V, and a conclusion is finally given in Section VI.

## II. RELATED WORK

### A. Fully-supervised Multi-organ Segmentation

Due to the high cost associated with manual organ delineation, researchers are exploring DL-based methods to achieve automatic MoS. For example, inspired by the fully convolutional network (FCN), Dou *et al.* [18] introduced a 3D segmentation network with a deep supervision mechanism and fulfilled the automatical segmentation of the liver and heart in radiology scans. Nevertheless, these methods may lead to suboptimal performance for complex MoS tasks owing to the diverse variations among organs and subjects. Therefore, more methods with coarse-to-fine strategy are then introduced for accuracy improvements. To illustrate, Hu *et al.* [19] used a convolutional neural network (CNN) to produce rough organ delineations which was then refined with a time-implicit multi-phase level-set algorithm. Wang *et al.* [20] presented a two-stage organ-attention network for abdominal organ segmentation to make predictions from three views (i.e., axial, sagittal, and coronal) which were statistically fused with the guide of the local structural similarity, thus producing a better overall segmentation with less irrelevant information. More recently, Ma *et al.* [21] used a coarse-to-fine segmentation network to optimize the segmentation of small- and large-size organs, respectively, achieving good accuracy in the segmentation of organs at risk (OARs) for NPC. However, despite the relatively better performance achieved by the aforementioned methods, they primarily focus on training deep models in a fully supervised way, which inevitably go through a significant performance drop in a more practical semi-supervised scenario with limited labeled data.

### B. Semi-supervised Medical Image Segmentation

Semi-supervised learning (SSL), which aims to harness limited labeled data and abundant unlabeled data to develop stronger models with higher performance, has recently garnered notable attention in the field of medical image segmentation. Existing SSL segmentation methods can be roughly partitioned into four categories: consistency regularization [6, 7, 22], proxy-label methods [23], generative models [24], and hybrid methods [25]. Among these strategies, consistency regularization is widely recognized as a popular approach that endeavors to maintain similarity in model outputs under diverse perturbations. The main focus lies in effectively mining more knowledge from unlabeled data to regularize the learning of deep models. Following this philosophy, numerous works have emerged to perform medical image segmentation in semi-supervised scenarios. For instance, Yu *et al.* [26] utilized the mean teacher [5], a mainstream regularization architecture, and introduced an uncertainty-aware scheme to encourage the student model to learn more reliable knowledge, thus obtaining higher accuracy in the segmentation of the 3D left atrium. Additionally, Luo *et al.* [27] employed a pyramid-prediction network to learn the information from the unlabeled data by constraining the difference between each pyramid prediction and their average. More recently, Basak *et al.* [28] proposed a new interpolation consistency training (ICT) algorithm to inspire consistent interpolation prediction between two unlabeled data and their corresponding segmentation maps. Generally, semi-supervised learning has been widely applied in medical image segmentation. However, the majority of approaches mainly focus on single ROI segmentation, and a notable decline in performance occurs when directly applying them to segment multiple ROIs. Therefore, the semi-supervised MoS remains a relatively new research field that needs further exploration.

*C. Class-imbalance Learning*

Class-imbalance learning has been extensively studied in fully-supervised tasks and can be categorized into three categories, i.e., re-weighting [14, 29], re-sampling [30], and meta-learning [31]. Re-weighting assigns a higher penalty to prediction errors of small targets, while re-sampling involves over-sampled data with fewer instances to obtain a balanced distribution. In a similar vein, meta-learning selects the class-balanced labeled data and utilizes a validation loss as the meta objective for updating the deep networks. Nonetheless, these three methods depend on accurate labels and are not applicable to semi-supervised tasks. Nowadays, several works have emerged in the literature to address class-imbalance learning in semi-supervised settings. For example, Huynh *et al.* [32] introduced an adaptive blended consistency loss into the perturbation-based semi-supervised network to adaptively adjust the target class distribution, overcoming data skew to some extent. Wei *et al.* [33] proposed a class-rebalancing self-training strategy that select pseudo labels of minority classes more frequently based on the class distribution. However, they ignore the further generation or utilization of the balanced data, leading to suboptimal performance. Different from them, our method not only generates unbiased subclasses from the original one but also further exploits the balanced subclass to regularize the main MoS task.

III. MATHEDOLOGY

The overview of the proposed BASIC is depicted in Fig. 1 which first generates the class-balanced subclasses and then performs the semi-supervised MoS with balanced subclass-based regularization and semantic-conflict penalty. Specifically, to produce balanced subclasses with abundant unbiased knowledge, a pre-trained backbone maps the labeled data into pixel-level semantic features. Meanwhile, a balanced clustering algorithm [34] is utilized to cluster the features with each organ, thus transforming the original labels into multiple balanced subclasses with nearly equal numbers of pixels. Subsequently, to accomplish the semi-supervised MoS, we design the whole segmentation network based on the mean teacher [5] architecture, of which both the labeled and unlabeled data are inputted into the teacher and student models. The student model serves as the target model for training, whereas the teacher model is updated with the exponential moving average (EMA) by the student model at each training step. The predictions generated by the teacher model provide additional supervision for the learning process of the student model. Following the philosophy of multi-task learning, both models, including an auxiliary subclass segmentation (SCS) subnetwork and a main MoS subnetwork, harness the same structure and are constructed by a shared encoder and two task-specific decoders. After gaining the corresponding outputs from two separated decoders, balanced subclass regularization is innovatively designed to further constrain the student predictions of MoS task with the teacher predictions of SCS task, thus transferring unbiased knowledge from the SCS subnetwork to the MoS subnetwork and alleviating the negative influence of the class-imbalance issue.

To simplify the description of our method, we provide the notations used throughout this paper beforehand. In our semi-supervised problem setting, the labeled set is represented as $D_L = \{(x_L^i, y_L^i)\}_{i=1}^{N}$ where $x_L^i \in R^{H \times W}$ represents the input image with height $H$ and width $W$, and $y_L^i \in \{0,1 \dots K\}^{H \times W}$ denotes the segmentation labels with $K$ total organ substructures (0 for background) to be segmented. The unlabeled set is defined as $D_U = \{x_U^i\}_{i=N+1}^{N+M}$ where $N \ll M$. More details will be provided in the subsequent subsections.

*A. Class-balanced Subclass Partition*

Considering the serious class imbalance problem caused by the large size differences among different organs, directly training a MoS model with original data may lead to an unsatisfactory performance for small organs. To address this, we design a class-balanced subclass partition strategy to first separate the original classes (also called parent classes) into several class-balanced subclasses with almost equal pixel numbers. Concretely, we adopt U-net [35] as the backbone and train it with the labeled set $D_L$ with a supervised segmentation loss, thus enabling it with the fundamental ability of feature extraction. To perform pixel clustering and generate balanced data, we omit the output layer in the pre-trained backbone and map the labeled image $x_L$ into pixel-level semantic features $F_L = \{f_i\}_{i \in [1,p]}$, where $p$ represents the total pixel number and $f_i \in R^{1 \times C}$ denotes the feature vector of the $i$-th pixel. Next, we aggregate feature vectors of the same class to create several clusters, which are subsequently regarded as subclasses. Notably, we adopt the balanced clustering [34], rather than other conventional clustering methods (e.g., k-means clustering [36]) to adjust the pixel number in each cluster based on the pixel proportions of the original classes. Thus, the larger/smaller targets are divided into more/fewer subclasses, resulting in multiple subclasses with nearly equal numbers of pixels. Once all the original classes have been re-divided, a new balanced subclass label $y_{L_{sub}} \in \{0, 1 \ldots K_{sub}\}^{H \times W}$ is obtained, where $K_{sub}$ is the total number of subclasses. Subsequently, the class-balanced labeled dataset $D_{L_{sub}} = \{(x_L^i, y_{L_{sub}}^i)\}_{i=1}^N$ is further utilized for following regularizations.

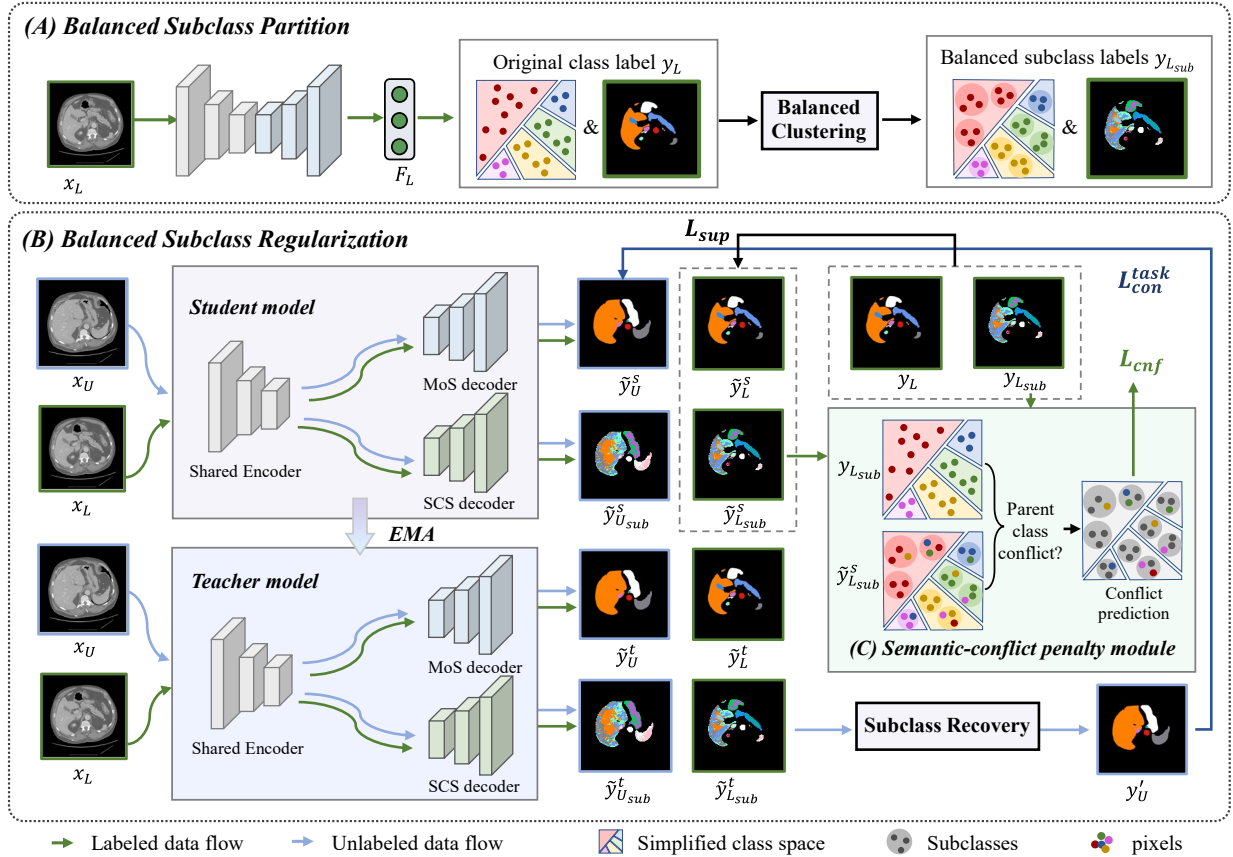

**Fig. 1.** Overview of the proposed BASIC network which first generates balanced subclass data with balanced clustering and then performs a balanced subclass regularization between the main MoS task and auxiliary SCS task. The whole network follows the architecture of mean teacher to accomplish the semi-supervised learning.

### B. Multi-task Learning within Mean Teacher Architecture

To achieve an effective semi-supervised MoS, there are two critical questions need to be addressed: (1) how to simultaneously utilize the limited labeled data $D_L$ and abundant unlabeled data $D_U$ to construct a robust semi-supervised framework and (2) how to incorporate the unbiased knowledge from the class-balanced data to enhance the accuracy of imbalanced targets in MoS. To tackle the first question, we draw inspiration from the notable performance of the mean teacher [5, 6] and avail it as our framework for semi-supervised learning. To address the second question, following the idea of multi-task learning, we devise a main MoS task and an auxiliary subclass segmentation (SCS) task where the two tasks are incorporated with a shared encoder and two task-specific decoders. The SCS task deeply excavates the unbiased knowledge in the class-balanced data $D_{L_{sub}}$ and we utilize its outputs to provide a balanced subclass regularization to the main MoS task, thus transferring the unbiased knowledge from the SCS subnetwork to the MoS network. Notably, the consistency constraints also further exploit the unlabeled data for reaching higher accuracy. More details will be described in the following subsections.

**Student Model:** We also employ the U-net [35] as the backbone for the main MoS task and the auxiliary SCS task. Notably, the encoder is shared by the two tasks while the parameters in the two task-specific decoders are different to fit different tasks. In this way, the encoder is enforced to capture crucial unbiased features with the optimization of the SCS network. Particularly, the shared encoder contains four down-sampling blocks, and the first three blocks are constructed by two 3×3 convolution layers followed by a batch normalization (BN) layer and a rectified linear unit (ReLU) activation function. Meanwhile, the strides of two convolutional layers in the block are set to 1 and 2, respectively, to gain the compressed feature representations. The fourth down-sampling block is equipped with two 3×3 convolutional layers with a stride of 1 and outputs the final feature map of the encoder. Besides, the two task-specific decoders share the same structures, each containing three up-sampling blocks and a task-specific head. The structure of the three up-sampling blocks is symmetrical to that of the first three down-sampling blocks in the encoder, except that the down-sampling operation is replaced with an up-sampling one to gradually restore the size of the feature map. Both the MoS head and SCS head apply the Softmax activation function.

Fed with a labeled image $x_L$ (unlabeled image $x_U$), the two subnetworks produce the MoS prediction $\tilde{y}_L^s$ ($\tilde{y}_U^s$) and SCS prediction $\tilde{y}_{L_{sub}}^s$ ($\tilde{y}_{U_{sub}}^s$) with the following formula:

$$\tilde{y}_L^s = f_{mos}(x_L; \theta_{mos}, \varepsilon), \tilde{y}_{L_{sub}}^s = f_{scs}(x_L; \theta_{scs}, \varepsilon), \quad (1)$$

$$\tilde{y}_U^s = f_{mos}(x_U; \theta_{mos}, \varepsilon), \tilde{y}_{U_{sub}}^s = f_{scs}(x_U; \theta_{scs}, \varepsilon), \quad (2)$$

where $f_{mos}$ and $f_{scs}$ denote the MoS and SCS subnetwork with corresponding parameters $\theta_{mos}$ and $\theta_{scs}$, respectively, and $\varepsilon$ represents the data perturbation in the student model.

**Teacher Model:** The teacher model follows the same architecture as the student model but their update strategies are different. The student model updates its parameters $\theta = \{\theta_{mos}, \theta_{scs}\}$ through the conventional gradient descent while the teacher model updates its parameters $\theta' = \{\theta'_{mos}, \theta'_{scs}\}$ by EMA. This process is formulated as:

$$\theta'_t = \alpha \theta'_{t-1} + (1-\alpha)\theta_t, \quad (3)$$

where $t$ represents the training step and $\alpha$ denotes the coefficient of EMA to control the updating rate.

Consistent with the student model, inputted with an image $x_L$ ($x_U$), the teacher model also outputs the MoS prediction $\tilde{y}_L^t$ ($y_U^t$) and SCS prediction $\tilde{y}_{L_{sub}}^t$ ($\tilde{y}_{U_{sub}}^t$) as follows:

$$\tilde{y}_L^t = f_{mos}(x_L; \theta'_{mos}, \varepsilon'), \tilde{y}_{L_{sub}}^t = f_{scs}(x_L; \theta'_{scs}, \varepsilon'), \quad (4)$$

$$\tilde{y}_U^t = f_{mos}(x_U; \theta'_{mos}, \varepsilon'), \tilde{y}_{U_{sub}}^t = f_{scs}(x_U; \theta'_{scs}, \varepsilon'), \quad (5)$$

where $\varepsilon'$ is the data perturbation in the teacher model. Then, the predictions made by the teacher model are utilized as the additional supervision for those of the student model.

### C. Balanced Subclass Regularization

The subclass labels are priorly subdivided from the original ones, so the main MoS and auxiliary SCS task theoretically maintain the same semantic information. Furthermore, unbiased knowledge in the SCS task, especially when dealing with class-imbalanced data, is crucial for enhancing the performance of the main MoS task, particularly on smaller targets. To this end, we propose a novel balanced subclass regularization via a task consistency loss, i.e., $L_{con}^{task}$, between the predictions of these two tasks, therefore introducing unbiased knowledge to the main MoS subnetwork. Specifically, as the affiliation relationship between the parent classes and subclasses is easy to acquire referring to the partition operation, we can map subclass prediction of teacher model $\tilde{y}_{U_{sub}}^t$ to parent class prediction $y_U'$ with negligible computational burden:

$$y_U' = M(\tilde{Y}_{U_{sub}}^t), \quad (6)$$

where the $M(\cdot)$ represents a mapping function.

To effectively transfer the unbiased knowledge in the SCS task to the MoS task, we can exploit the task consistency loss, i.e., $L_{con}^{task}$ to supervise the student prediction of MoS task with the mapped teacher prediction $y_U'$ of SCS task. $L_{con}^{task}$ can be expressed as below:

$$L_{con}^{task} = L_{ce}(y_U', \tilde{Y}_U^s) + L_{dice}(y_U', \tilde{Y}_U^s), \quad (7)$$

where $L_{ce}$ and $L_{dice}$ represent the commonly-used cross-entropy (CE) loss and Dice loss, respectively.

### D. Semantic-conflict Penalty Mechanism

As stated in Section III.B, the parent class prediction recovered from subclass prediction is used to achieve balanced subclass regularization, therefore the accurate subclass predictions are critical for unbiased knowledge transferring to the main MoS task. Considering the important hierarchical semantic relationship between the parent classes and subclasses (i.e., subclasses derived from the same parent class share similar semantic information), wrong SCS predictions can be divided into two types: wrong but unconflicting SCS prediction which segments a pixel into another incorrect subclass of the correct parent class, and wrong and conflicting SCS prediction which segments a pixel into an incorrect subclass of the incorrect parent class. Even though two kinds of wrong predictions are constrained by supervised loss on labeled data, the conflicting predictions pose more negative impacts owing to their damage to following balanced subclass regularization with inconsistent semantic information. To enforce the network to pay more attention to such conflicting predictions, we present a semantic-conflict penalty mechanism for providing heavier punishments to the conflicting SCS predictions with wrong parent classes. As shown in Fig.1(C), we check the consistency between the SCS predictions and their parent classes in the ground truth and impose the semantic-conflict pixels with a higher penalty through the following conflict loss:

$$L_{cnf} = \mathbb{1}\left[M\left(\tilde{y}_{L_{sub}}^s(i,j)\right) \neq M\left(y_{L_{sub}}(i,j)\right)\right] \times L_{ce}(\tilde{y}_{L_{sub}}^s, y_{L_{sub}}), \quad (8)$$

where $(i,j)$ indicates the pixel coordinate and $\mathbb{1}[\cdot]$ represents a binary indicator.

In this way, the network can better ensure semantic consistency between the subclasses belonging to the same parent classes, thus generating a more precise supervision of the main MoS task.

### E. Objective Functions

The objective function of the proposed method is composed of four parts: (1) supervised segmentation loss ($L_{sup}$) to measure the disparity between the ground truth and prediction results of the student, (2) model consistency loss ($L_{con}^{model}$) between the prediction results of the student and teacher, (3) task consistency loss ($L_{con}^{task}$) between the predictions of the MoS task and SCS task, and (4) conflict loss ($L_{cnf}$) to correct the semantic-conflict SCS predictions.

Since the labeled data $x_L$ has a manually delineated ground truth $y_L$ and corresponding subclass label $y_{L_{sub}}$, we conduct a supervised constraint to narrow the difference between the predictions, i.e., $\tilde{Y}_L^s$ and $\tilde{Y}_{L_{sub}}^s$, and their corresponding targets as below:

$$L_{sup} = L_{seg}(y_L, \tilde{y}_L^s) + \mu L_{seg}(y_{L_{sub}}, \tilde{y}_{L_{sub}}^s), \quad (9)$$

where $L_{seg}$ also incorporates CE loss and Dice loss equally and $\mu$ is a weighted term.

Besides, following the design of the mean teacher [5], we introduce the model consistency loss $L_{con}^{model}$ to force the prediction of an unlabeled input $x_U$ from the student to keep similar to that from the teacher, which is formulated as follows:

$$L_{con}^{model} = L_{mse}(\tilde{y}_U^s, \tilde{y}_U^t) + L_{mse}(\tilde{y}_{U_{sub}}^s, \tilde{y}_{U_{sub}}^t), \quad (10)$$

where $L_{mse}$ means a mean-square error (MSE).

Consequently, the total loss function $L_{total}$ can be formulated as:

$$L_{total} = L_{sup} + \lambda_1 L_{con}^{model} + \lambda_2 L_{con}^{task} + \lambda_3 L_{cnf}, \quad (11)$$

where $\lambda_i$ ($i\epsilon\{1,2,3\}$) are the hyper-parameters to balance the sub-terms.

## IV. EXPERIMENTS AND RESULTS

### A. Dataset and Processing

**WORD Dataset:** The WORD dataset is a large-scale Whole abdominal Organ Dataset [16] with 150 computed tomography (CT) volumes. There are 16 organ annotations: liver (Liv), spleen (Spl), left kidney (L.kid), right kidney (R.kid), stomach (Sto), gallbladder (Gall), esophagus (Eso), pancreas (Pan), duodenum (Duo), colon (Col), intestine (Int), adrenal (Adr), rectum (Rec), bladder (Bla), left head of the femur (L.hf), and right head of the femur (R.hf). We follow the official partitions which utilize 100, 20, and 30 samples as training, validation, and testing set, respectively.

**MICCAI FLARE 2022 dataset:** The MICCAI FLARE 2022 dataset is a subset of the abdomen CT image segmentation Flare challenge [17] which randomly select 135 patient samples. Besides nine shared organs with the WORD dataset, i.e., Liv, L.kid, R.kid, Spl, Pan, Sto, Gall, Duo, and Eso, there are four specific organs needed to be segmented: aorta (Aor), inferior vena cava (Ivc), right adrenal gland (Rag), and left adrenal gland (Lag). We randomly select 100, 10, and 25 samples as training, validation, and testing set, respectively.

All volumes are sliced along axial direction into 2D images with size of 512×512. For the two datasets, in the training set, we divide the labeled set and the unlabeled set as $n/m$ to simulate the semi-supervised setting, where $n$ and $m$ are the numbers of labeled and unlabeled samples. In addition, we use random rotation and Gaussian noising for data augmentation to improve the model generalization and prevent the over-fitting problem.

### B. Experimental Settings and Evaluation Metrics

**Experimental Settings:** The proposed BASIC network is implemented with the PyTorch framework and trained on a single NVIDIA GeForce RTX 3090 GPU with a total memory of 24GB. Adam optimizer is employed to train the whole model for 30000 iterations with a learning rate of 1e-2 and batch size of 16. As for the hyper-parameters, $\mu$ in Eq. (9) is set as 1 and $\lambda_2$ in Eq. (11) is set as 0.5. Following [5], $\alpha$ in Eq. (3) is set as 0.99 and $\lambda_1$ in Eq. (11) is set as $0.1 \times e^{(-5(1-\delta/\delta_{max})^2)}$, where $\delta$ and $\delta_{max}$ denote the current training step and total training steps. $\lambda_3$ is empirically set as 1. Besides, the SCS decoder is utilized to help the training of the

TABLE I

QUANTITATIVE RESULTS OF SOTA METHODS ON WORD DATASET WHEN N=5, 10, AND 15, RESPECTIVELY. THE BEST RESULT OF EACH INDEX IS MARKED IN **BOLD** WHILE THE SECOND-BEST ONE IS UNDERLINED.

| Method | Liv | Spl | L.kid | R.kid | Sto | Gall | Eso | Pan | Duo | Col | Int | Adr | Rec | Bla | L.hf | R.hf | Avg |
|---|---|---|---|---|---|---|---|---|---|---|---|---|---|---|---|---|---|
| Total Number of Labeled data: n=5 |||||||||||||||||
| U-net | 84.11 | 64.36 | 80.52 | 82.04 | 61.17 | 36.67 | 50.50 | 43.87 | 35.72 | 66.59 | 63.55 | 44.04 | 59.75 | 78.99 | 88.80 | 88.38 | 64.32 |
| MT | 92.52 | 75.29 | 81.28 | 83.69 | 46.64 | 39.98 | 57.84 | 48.56 | 34.12 | 68.07 | 68.07 | 38.00 | 63.50 | 82.25 | 87.78 | **90.94** | 66.16 |
| UAMT | 92.67 | 72.69 | 79.93 | 82.22 | 55.16 | 49.99 | 57.26 | 53.56 | 33.78 | 67.82 | 68.92 | 35.69 | 59.39 | 78.72 | 80.86 | 89.27 | 66.12 |
| ICT | 92.80 | 70.87 | 83.54 | 83.92 | 60.29 | 54.27 | 56.50 | 50.16 | 40.44 | 69.03 | 69.49 | 42.42 | 62.99 | 83.31 | 88.46 | 88.79 | 68.58 |
| URPC | 91.13 | 76.98 | 83.90 | 83.92 | 59.22 | 43.43 | 58.27 | 50.43 | 36.34 | 63.82 | 68.22 | 42.09 | 63.99 | 82.94 | 85.06 | 89.67 | 67.46 |
| EVIL | 92.16 | 76.69 | 82.81 | 82.37 | 54.93 | 53.48 | 56.08 | 58.24 | 34.51 | 66.88 | 69.96 | 47.75 | 64.21 | 83.14 | 88.77 | 88.07 | 68.75 |
| Ours | **94.17** | **84.69** | **85.55** | **86.32** | **73.52** | **54.50** | **62.57** | **59.75** | **43.18** | **70.63** | **71.03** | **45.29** | 63.65 | **90.05** | **88.90** | 90.83 | **72.79** |
| Total Number of Labeled data: n=10 |||||||||||||||||
| U-net | 90.21 | 69.13 | 84.13 | 84.19 | 73.58 | 58.47 | 54.16 | 45.56 | 39.52 | 70.31 | 68.42 | 50.23 | 63.43 | 83.47 | 90.08 | 86.52 | 69.46 |
| MT | 93.39 | 82.37 | 82.99 | 83.42 | 69.79 | 55.33 | 64.01 | 56.32 | 40.33 | 70.68 | 64.70 | 46.86 | 62.16 | 83.36 | 87.43 | 90.32 | 70.84 |
| UAMT | 92.76 | 79.65 | 81.80 | 83.95 | 75.28 | 41.62 | 65.94 | 57.12 | 42.14 | 71.08 | 71.60 | 43.24 | 62.49 | 79.64 | 89.11 | 90.43 | 70.49 |
| ICT | 91.91 | 74.59 | 83.20 | 87.36 | 78.04 | 63.05 | 68.27 | 53.66 | 40.15 | 70.14 | 67.00 | 48.77 | 54.27 | 89.56 | 90.38 | 91.71 | 72.00 |
| URPC | 93.61 | 80.77 | 85.53 | 85.81 | 73.75 | 66.43 | 64.65 | 57.87 | 45.87 | 70.72 | 72.10 | 44.66 | 68.53 | 85.05 | 88.23 | 87.52 | 73.19 |
| EVIL | 94.26 | 79.11 | 85.00 | 87.23 | 78.56 | **71.68** | 62.00 | 62.75 | 47.68 | 74.88 | 74.79 | 53.71 | 71.32 | 88.40 | 87.78 | 89.07 | 75.51 |
| Ours | **94.70** | **88.14** | **88.19** | **88.58** | **80.20** | 71.32 | **67.72** | **64.47** | **49.27** | **75.82** | **77.18** | **49.87** | 66.68 | **90.51** | **91.98** | **92.63** | **77.33** |
| Total Number of Labeled data: n=15 |||||||||||||||||
| U-net | 94.02 | 91.73 | 91.17 | 93.88 | 68.73 | 56.65 | 68.19 | 69.26 | 53.52 | 75.79 | 76.05 | 52.83 | 66.53 | 64.31 | 85.94 | 89.12 | 74.86 |
| MT | **94.20** | 89.98 | 91.12 | 90.12 | 71.62 | 62.88 | 67.42 | 65.20 | 55.11 | 75.38 | 74.48 | 55.48 | 68.84 | 87.03 | 86.52 | 89.05 | 76.53 |
| UAMT | 90.59 | 89.88 | 90.45 | 90.95 | 67.13 | 48.71 | 68.86 | 60.14 | 48.91 | 74.56 | 71.17 | 52.23 | 67.01 | 82.34 | 86.96 | 89.91 | 73.74 |
| ICT | 94.19 | 91.09 | 89.45 | 93.37 | 76.39 | 59.10 | 67.94 | 68.29 | 52.53 | 76.53 | 77.22 | 53.28 | **71.28** | 90.21 | 81.25 | **90.23** | 77.02 |
| URPC | 93.96 | 88.91 | 92.03 | 92.84 | 77.02 | 59.15 | 71.29 | 68.27 | 55.70 | 77.61 | 78.00 | 47.88 | 69.70 | 87.47 | 89.44 | 89.85 | 77.45 |
| EVIL | 91.80 | 91.72 | 90.48 | 92.92 | 73.45 | 60.22 | 70.20 | 71.51 | 56.55 | 75.75 | 77.22 | 56.45 | 70.82 | 90.43 | 80.21 | 88.13 | 77.37 |
| Ours | 93.13 | **91.81** | **93.88** | **94.90** | **80.26** | **66.70** | **72.39** | **72.42** | **58.09** | **78.01** | **79.13** | **56.71** | 68.95 | **90.82** | 88.15 | 89.01 | **79.65** |

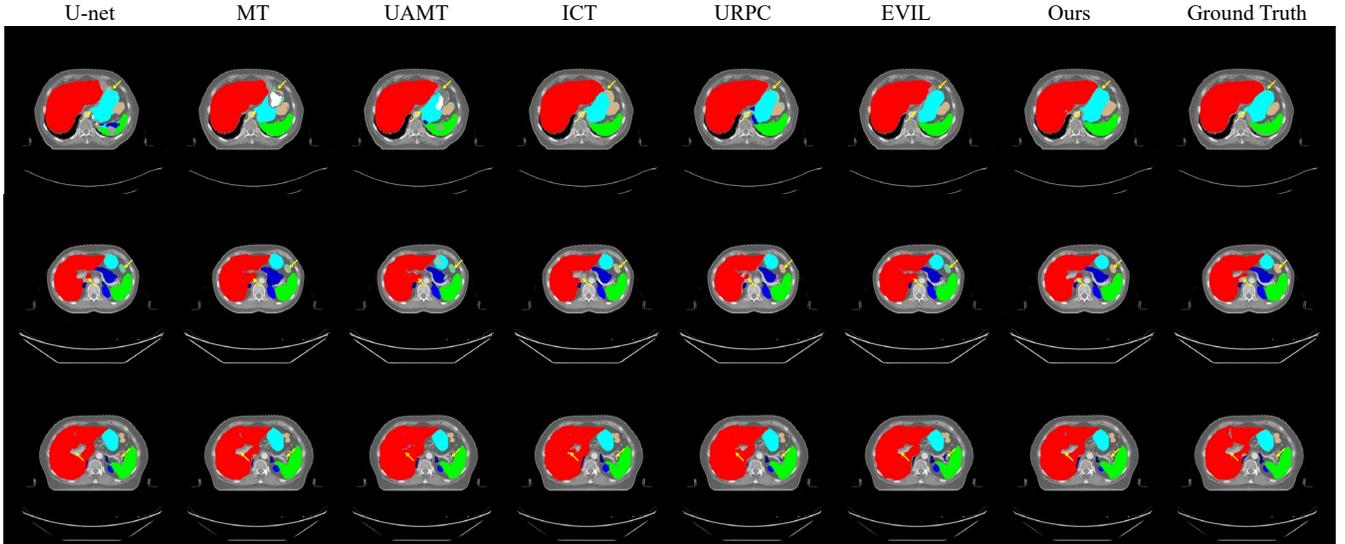

Fig. 2. Visualization comparisons with SOTA models on WORD dataset. From top to bottom, n=5, 10, and 15, respectively.

main MoS network during the training stage while in the testing stage, only the shared encoder and the MoS decoder are reserved for the final segmentation. Moreover, the teacher model is chosen as the final prediction model for its better stability and generalization.

*Evaluation Metrics:* To quantitatively measure the performance, we utilize a commonly-used Dice coefficient ($Dice = \frac{2*|A \cap B|}{|A|+|B|}$) [37] to measure the overlapping between the prediction and the ground truth. Higher Dice represents better performance. Besides, we give corresponding visualizations to intuitively display the comparison of our method and others.

### C. Comparison with the State-of-the-art Methods

To demonstrate the superior performance of our BASIC in semi-supervised MoS, we compare it with multiple methods, i.e., U-net (2015) [24], mean teacher (MT, 2017) [5], uncertainty aware mean teacher (UAMT, 2019) [26], interpolation consistency training (ICT, 2022) [28], uncertainty rectified pyramid consistency (URPC, 2022) [27], and evidential inference learning (EVIL, 2023) [38].

*Comparison on WORD dataset:* The quantitative results on the WORD dataset are reported in Table I where the proposed gains the best overall performance for all the data partitions. Concretely, confronting the most extreme setting, i.e., $n$=5, MT and UAMT gain relatively lower results, i.e., 66.16% and 66.12% for the average Dice. Besides, EVIL obtains the second-best accuracy, i.e., 68.75% average Dice, and performs well on Adr (47.75%) and Rec (64.21%). However, on some other organs, the performance of EVIL is unsatisfactory even compared to ICT, i.e., Duo (↓5.93%), Sto (↓5.36%), and Col (↓2.15%). Compared to EVIL, our BASIC gains notable enhancements for Spl (↑8.00%), Sto (↑18.59%), and Duo (↑8.67%). When $n$ increases to 10 and 20, all the methods achieve higher accuracy with stronger support from labeled data. Under these two partitions, we can also observe that our BASIC maintains its leading performance. Besides, the intuitive visualizations are displayed in Fig.2 where our method can accurately segment the targets with the least fault segmentation.

TABLE II
QUANTITATIVE RESULTS OF SOTA METHODS ON FLARE DATASET WHEN N=5 AND 10, RESPECTIVELY. THE BEST RESULT OF EACH INDEX IS MARKED IN **BOLD** WHILE THE SECOND-BEST ONE IS UNDERLINED.

| Method | Liv | Spl | L.kid | R.kid | Sto | Gall | Eso | Pan | Duo | Col | Int | Adr | Rec | Avg |
|---|---|---|---|---|---|---|---|---|---|---|---|---|---|---|
| Total Number of Labeled data: $n$=5 ||||||||||||||
| U-net | 88.95 | 57.51 | 70.83 | 45.48 | 80.93 | 68.79 | 58.83 | 58.33 | 62.00 | 62.49 | 67.86 | 39.10 | 64.36 | 63.50 |
| MT | 93.56 | 77.72 | 80.46 | 52.52 | 87.24 | 76.05 | 55.71 | 52.59 | 41.35 | 68.82 | 71.90 | 36.59 | 83.08 | 67.51 |
| UAMT | 91.51 | 73.22 | 85.76 | 45.47 | 86.44 | 74.76 | 62.98 | 58.79 | 60.79 | 67.22 | 66.15 | 34.71 | 77.08 | 68.07 |
| ICT | 92.21 | 73.83 | 89.35 | 51.26 | 86.70 | 76.16 | 63.00 | 60.82 | 57.58 | 68.80 | **74.51** | 38.91 | 77.11 | 70.02 |
| URPC | 92.39 | 73.46 | 86.33 | 48.44 | 88.27 | 74.92 | 61.89 | **62.16** | 68.29 | 69.50 | 63.24 | 41.29 | 78.91 | 69.93 |
| EVIL | 93.98 | 83.38 | 88.89 | 50.13 | 85.15 | 76.41 | 67.82 | 59.54 | 66.59 | 69.26 | 70.03 | 41.60 | 85.99 | 72.21 |
| Ours | **96.30** | **94.38** | **95.14** | **62.21** | **93.58** | **82.66** | **71.04** | 61.87 | **73.00** | **72.28** | 74.47 | **50.16** | **91.26** | **78.34** |
| Total Number of Labeled data: $n$=10 ||||||||||||||
| U-net | 94.41 | 93.14 | 92.77 | 56.78 | 91.54 | 80.97 | 70.02 | 70.63 | 62.23 | 72.11 | 80.00 | 51.15 | 92.84 | 77.58 |
| MT | 95.92 | 92.85 | 95.70 | 61.56 | 91.65 | 83.31 | 68.53 | 62.59 | 67.94 | 73.95 | 80.18 | 55.33 | 92.32 | 78.60 |
| UAMT | 95.37 | 95.86 | 94.25 | 61.70 | 92.01 | 82.35 | 74.15 | **71.48** | 66.78 | 71.25 | 80.40 | 56.40 | 93.67 | 79.67 |
| ICT | 96.43 | 95.24 | 94.92 | 61.55 | 92.90 | 82.36 | 74.74 | 69.63 | 69.30 | 76.82 | 84.62 | 56.97 | 93.31 | 80.68 |
| URPC | 96.86 | 94.41 | 95.35 | 64.45 | 92.41 | 84.54 | **76.23** | 69.92 | 69.34 | 75.93 | 82.92 | 58.56 | 92.62 | 81.04 |
| EVIL | 96.52 | 94.70 | 96.47 | 64.86 | 93.93 | 85.45 | 71.09 | 69.72 | 78.19 | **79.58** | 81.45 | 58.93 | 90.66 | 81.66 |
| Ours | **97.45** | **96.09** | **97.06** | **69.21** | **94.57** | **88.48** | 73.55 | 70.74 | **82.34** | 78.47 | **87.36** | **62.79** | **94.41** | **84.04** |

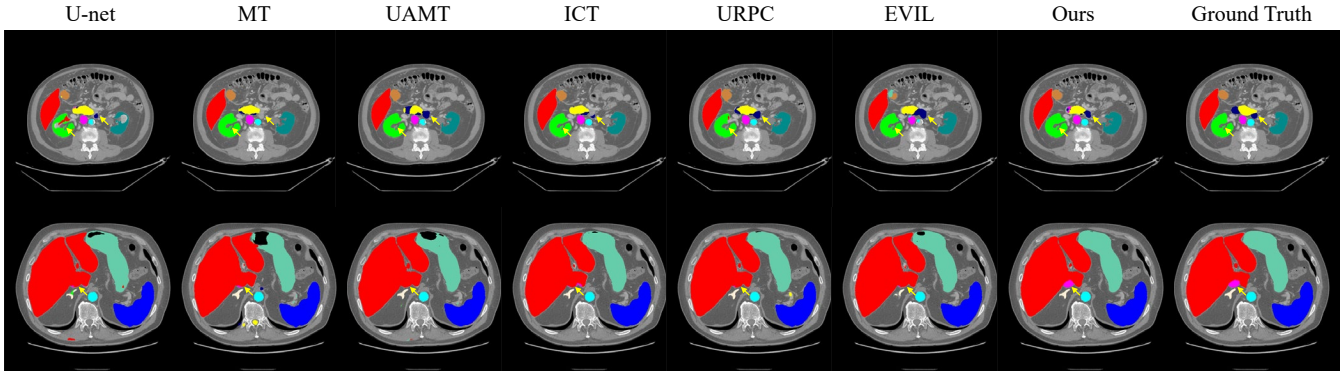

**Fig. 3.** Visualization comparisons with SOTA models on FLARE dataset. From top to bottom, n=5 and 10, respectively.

TABLE III
ABLATION STUDIES OF KEY COMPONENTS ON WORD AND FLARE DATASETS WHEN N=5. THE BEST RESULT OF EACH INDEX IS MARKED IN **BOLD** WHILE THE SECOND-BEST ONE IS <u>UNDERLINED</u>.

| Variants | WORD | | | | | | | FLARE | | | | | | |
|---|---|---|---|---|---|---|---|---|---|---|---|---|---|---|
| | Liv | R.kid | Sto | Int | Rec | R.hf | Avg | Liv | R.kid | Sto | Eso | Int | Rec | Avg |
| (A) | 84.11 | 82.04 | 61.17 | 63.55 | 59.75 | 88.38 | 64.32 | 88.95 | 45.48 | 80.93 | 58.83 | 67.86 | 64.36 | 63.50 |
| (B) | 92.52 | 83.69 | 46.64 | 68.07 | <u>63.50</u> | <u>90.94</u> | 66.16 | 93.56 | 52.52 | 87.24 | 55.71 | 71.90 | 83.08 | 67.51 |
| (C) | <u>93.88</u> | 83.49 | 57.55 | <u>68.97</u> | 59.67 | 89.70 | 68.71 | 92.53 | 55.96 | 89.46 | 62.48 | 70.76 | 79.19 | 71.87 |
| (D) | 92.02 | <u>85.88</u> | <u>72.24</u> | 70.51 | 58.79 | 88.92 | <u>71.82</u> | 94.51 | <u>58.51</u> | 91.12 | <u>66.56</u> | **74.51** | <u>87.81</u> | 76.81 |
| (E) | 92.25 | 85.39 | 58.32 | 68.29 | 58.18 | 89.73 | 69.51 | <u>94.57</u> | 55.91 | 89.90 | 61.77 | 72.20 | 86.50 | 73.32 |
| (F) | **94.17** | **86.32** | **73.52** | **71.03** | **63.65** | **90.83** | **72.79** | **96.30** | **62.21** | **93.58** | **71.04** | <u>74.47</u> | **91.26** | **78.34** |

*Comparison on the MICCAI FLARE 2022 Dataset:* We provide the quantitative results on the MICCAI FLARE 2022 dataset in Table II. As seen, when $n$=5, the proposed method achieves the best overall performance (78.34% average Dice) and outperforms all comparative methods in terms of all organs, except for the Int which is slightly lower (↓0.04%) than ICT. Compared to the second-best EVIL, our proposed also significantly surpasses it by 6.13%. Additionally, concentrating on the small organs, our BASIC further promotes the segmentation accuracy notably, especially for Duo (↑6.41%), Adr (↑8.56%), and Rec (↑5.27%). Such superiorities are also clearly observed when there are 10 labeled data, i.e., $n$=10. More intuitively, Fig. 3 shows the visual results from where we can see that our BASIC also gains the best performance, especially for the tiny structures.

Based on the above results, we can conclude two points: (1) the notable performance enhancements are accomplished by our BASIC for all data partitions, sufficiently validating the effectiveness of BASIC on semi-supervised segmentation; (2) BASIC can successfully relieve the class imbalance issue in MoS and reach better performance on small organs.

### D. Ablation Studies

In this section, we conduct several ablation experiments on both WORD and FLARE datasets ($n$=5) to investigate the effectiveness of the key components in the proposed BASIC. The experimental arrangements can be summarized as: (A) U-net only trained with labeled data, (B) mean teacher with traditional model consistency loss (MT + $L_{con}^{model}$) as the backbone (MT), (C) MT + $L_{con}^{model}$ + balanced SCS task (bSCS), (D) MT + $L_{con}^{model}$ + bSCS + task consistency loss ($L_{con}^{task}$), (E) MT + $L_{con}^{model}$ + imbalanced SCS task + $L_{con}^{task}$ +semantic-conflict penalty loss ($L_{cnf}$) (proposed-K-means), (F) MT + $L_{con}^{model}$ + bSCS + $L_{con}^{task}$ +$L_{cnf}$ (Proposed). The quantitative results are displayed in Table III. Notably, we only report the respective results of six organs for page limits but the average results come from averaging the accuracy of all organs.

*Contribution of the Balanced Subclass Segmentation (SCS) Task:* To explore the contribution of balanced SCS task, we can compare the results of (B) and (C) where the balanced SCS task further enhance the overall accuracy from 66.16% to 68.71% for WORD dataset and from 67.51% to 71.87% for FLARE dataset, respectively, which verifying the positive influence of balanced SCS task for introducing the unbiased information to the main MoS task.

*Contribution of the Task Consistency Loss:* To validate the contribution of $L_{con}^{task}$, we can see the results of (C) and (D), with the help of additional task consistency constraint, the segmentation performance is further promoted by 3.11% for the WORD dataset and 4.94% for the FLARE dataset. Notably, the improvements gaining by introducing task consistency loss are the most significant among all key components, which strongly proves its effectiveness in transferring unbiased knowledge to the MoS subnetwork.

*Contribution of Semantic-conflict Penalty Mechanism:* We can directly compare the results of (D) and (F) to investigate the contribution of the semantic-conflict penalty mechanism. As seen, it effectively enhances the Dice from 71.82% to 72.79% for the WORD dataset and from 76.81% to 78.34% for the FLARE dataset, respectively, verifying its contribution to maintaining the semantic consistency between subclasses and their corresponding parent classes.

*Impact of Different Clustering Strategies:* We additionally design an exploration experiment to investigate the impact of different clustering strategies when generating subclasses. Concretely, (E) utilizes a traditional k-means clustering [36] while (F), i.e., the proposed method, exploits the balanced clustering [34] to produce subclasses from the original classes. Compared with (E), (F) gains the higher average Dice as well as better respective accuracy on all organs, verifying that balanced subclasses can effectively provide unbiased information and finally promote the segmentation results.

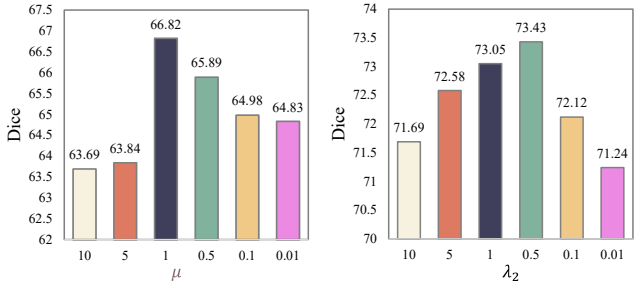

**Fig. 4.** The effect of different values of hyper-parameter $\mu$ (Left) and $\lambda_2$ (Right) on the validation set of WORD dataset.

***Selections of Hyper-parameters:*** To select the optimal hyper-parameters, i.e., $\mu$ and $\lambda_2$, to reach the optimal results, we conduct the following exploration experiments on the validation set of WORD dataset when $n$ =5. To find the optimal selection of the weighted term $\mu$ supervised segmentation loss $L_{sup}$, we utilize the U-net to fulfill the balanced SCS task as well as the main MoS task in a multi-task framework and adjust the value of $\mu$ gradually. For the hyper-parameter $\lambda_2$ of task consistency loss, we exploit variant (D) and explore the performance difference with diverse values of $\lambda_2$. All the experimental results are displayed in Fig. 4. As observed, when $\mu$ and $\lambda_2$ are set as 1 and 0.5, respectively, the optimal performance is gained. Therefore, we keep these settings in all of our experiments.

## V. DISCUSSION

In this paper, to alleviate the class-imbalance problem in semi-supervised multi-organ segmentation (MoS), we present a novel semi-supervised MoS network, i.e., BASIC, equipped with balanced subclass regularization and semantic-conflict penalty mechanism. The existing class-rebalance strategies in fully supervised tasks, e.g., re-weighting and re-sampling, mainly exploit the accurate labels to correct the biased predictions that are not applicable to the unlabeled data in the semi-supervised MoS. In this regard, we priorly generate unbiased subclasses from the original classes and design a novel auxiliary subclass segmentation (SCS) task to learn the important class-unbiased knowledge for the main MoS task These two tasks are fulfilled with a shared encoder and two respective decoders within a multi-task learning framework. To accomplish the semi-supervised learning, we construct a mean teacher architecture where the SCS and MoS subnetworks are involved in both the student and the teacher model. Then, a balanced subclass regularization is designed to utilize the predictions of the SCS task to supervise that of the MoS task, thus promoting an effective transfer of unbiased information. Finally, we design a semantic-conflict penalty mechanism to better maintain the semantic similarity between the subclasses and their corresponding original classes.

To verify the feasibility and generalizability of the proposed method, we performed the experiments on two public abdominal organ segmentation datasets, i.e., the sixteen-class WORD dataset and the thirteen-class MICCAI Flare 2022 dataset. In Section IV.C, we display the detailed qualitative results in Table I and Table II from which the proposed method not only performs the superior segmentation accuracy compared to all other SOTAs under all data partitions but also gains higher accuracy in terms of small organs which demonstrates its effectiveness on alleviating class imbalance issue. Visualizations in Fig.2 and Fig.3 indicate that the proposed BASIC also achieves the best results with the least wrong segmentation. To explore the contributions of the key components, we conduct several ablation studies in Section IV.C where the results in Table III verify the respective effectiveness of the key parts. Besides, we perform a hyper-parameter selection experiment and the results in Fig.4 validate the optimal choice of the two hyper-parameters.

Despite the notable performance enhancement achieved, our current work has the following limitations. First, although the semi-supervised strategy relieves the dependency on annotated data to some extent, labeled data is still required to guide the feature extraction which limits its practical applications. Consequently, constructing a robust unsupervised framework that only utilizes unlabeled data to achieve high-quality automatical segmentation is a worth exploring topic. Second, there are multiple modalities [39, 40], i.e., CT, positron emission tomography (PET), and Magnetic Resonance Imaging (MRI), with abundant complementary information which is also crucial for medical image segmentation. Therefore, we will explore how to effectively utilize multi-modal data in semi-supervised MoS to enhance the accuracy in the future.

## VI. CONCLUSION

In this paper, we present BASIC, a novel semi-supervised network with balanced subclass regularization and semantic-conflict penalty mechanism, to tackle the issue of class imbalance in semi-supervised MoS task. Based on the subclasses generated from their original (parent) classes, we introduce an auxiliary subclass segmentation (SCS) task and design a novel balanced subclass regularization to provide constraints to the main MoS task, thus effectively transferring the unbiased knowledge to the MoS subnetwork and relieving the class-imbalance problem. Furthermore, to leverage the semantic similarity between the subclasses and their parent classes, we introduce a semantic-conflict penalty mechanism to give heavier punishment to conflicting SCS predictions with the wrong parent class, thus reaching a more precise constraint to the main MoS predictions. Moreover, experimental results on the publicly available WORD dataset and MICCAI FLARE 2022 dataset have demonstrated the superior performance of our method.